%% file: main.tex
\documentclass[letterpaper]{article}
\usepackage{uai2019}
\usepackage[margin=1in]{geometry}
\usepackage{caption}
\usepackage{subcaption}

\newif\ifcomments
\commentstrue

\usepackage[utf8]{inputenc} 

\usepackage{times}
\usepackage{graphicx}
\usepackage{pdfcomment}
\usepackage{booktabs} 
\usepackage{hyperref}
\usepackage{natbib}
\usepackage[algo2e,ruled]{algorithm2e}

\usepackage{color}
\usepackage[dvipsnames]{xcolor}
\usepackage{amsmath,amsthm,verbatim,amssymb,amsfonts,amscd,mathrsfs,mathtools}
\newtheorem{prop}{Proposition}
\usepackage{multirow}
\usepackage{makecell}  
\usepackage{hhline}  
\usepackage{url}
\usepackage{physics}

\usepackage[UKenglish]{babel}

\makeatletter
\DeclareRobustCommand*\textsubscript[1]{%
  \@textsubscript{\selectfont#1}}
\def\@textsubscript#1{%
  {\m@th\ensuremath{_{\mbox{\fontsize\sf@size\z@#1}}}}}
\makeatother

\ifcomments\newcommand{\comments}[1]{#1}\else\newcommand{\comments}[1]{}\fi
\definecolor{clrgp}{rgb}{.9,0,.9}

\newcommand\blfootnote[1]{%
  \begingroup
  \renewcommand\thefootnote{}\footnote{#1}%
  \addtocounter{footnote}{-1}%
  \endgroup
}

\input{macros}

\renewcommand{\paragraph}[1]{\vspace{-0.5ex}\textbf{#1}}

\usepackage{color}
\definecolor{dark-blue}{rgb}{0.15,0.15,0.4}
\definecolor{medium-blue}{rgb}{0,0,0.5}
\hypersetup{
   colorlinks, linkcolor={dark-blue},
   citecolor={dark-blue}, urlcolor={medium-blue}
}

\renewcommand{\cite}[1]{\citep{#1}}

\title{Fenton-Wilkinson Order Statistics and German Tanks:\\A Case Study of an Orienteering Relay Race\\
}

\author{Joonas~P\"a\"akk\"onen, Ph.D.}

\begin{document}
  \maketitle

\begin{abstract}
Ordinal regression falls between discrete-valued classification and continuous-valued regression. Ordinal target variables can be associated with ranked random variables. These random variables are known as order statistics and they are closely related to ordinal regression. However, the challenge of using order statistics for ordinal regression prediction is finding a suitable parent distribution. In this work, we provide a case study of a real-world orienteering relay race by viewing it as a random process. For this process, we show that accurate order statistical ordinal regression predictions of final team rankings, or \emph{places}, can be obtained by assuming a lognormal distribution of individual leg times. Moreover, we apply Fenton-Wilkinson approximations to intermediate changeover times alongside an estimator for the total number of teams as in the notorious German tank problem. The purpose of this work is, in part, to spark interest in studying the applicability of order statistics in ordinal regression problems.\blfootnote{\today. Email: pkknen@yahoo.com.}
\end{abstract}

  \input{introduction}
  \input{method}
  \input{results}
  \input{discussion}
  \bibliography{bibliography}
  \bibliographystyle{apalike}

\end{document}

%% file: macros.tex
\newif\ifboldmatrix
\boldmatrixfalse
\ifboldmatrix\newcommand{\boldmatrix}[1]{\mathbf{#1}}\else\newcommand{\boldmatrix}[1]{#1}\fi
\newcommand{\transpose}{\ensuremath{^\top}}

\newcommand{\x}{\ensuremath{\mathbf{x}}}

\newcommand{\bk}{\ensuremath{\mathbf{k}}}

\newcommand{\by}{\ensuremath{\mathbf{y}}}

\newcommand{\A}{\ensuremath{\boldmatrix{A}}}

\newcommand{\K}{\ensuremath{\boldmatrix{K}}}
\renewcommand{\bk}{\ensuremath{\mathbf{k}}}

\newcommand{\y}{\ensuremath{\mathbf{y}}}

%% file: introduction.tex
\section{INTRODUCTION}
Machine learning includes various classification, regression, clustering and dimensionality reduction methods. In classification models, the target is a discrete class, while in regression, the target is typically a continuous variable. \emph{Ordinal regression}, also known as \emph{ordinal classification}, is regression with a target that is discrete and ordered. Ordinal classification can be thus considered to be a hybrid of classification and regression.

Typical applications of ordinal classification include age estimation with an integer-valued target, advertising systems, recommender systems, and movie ratings on a scale from $1$ to $5$. For further insight into recent developments in machine learning, including ordinal regression, the reader is kindly directed to \cite{Guti2016,Cao2019,Vargas2019,Lu2019,Gambella2019} and the references therein.

In this work, we perform a case study of ordinal regression on the ranks of a sorted sum of random variables corresponding to the duration of an orienteering relay viewed as a random process with a large number of subsamples of changeover times. We compare three widely-used regression schemes to our original method that is based on manipulations of certain properties of ordered random variables, otherwise known as \emph{order statistics}.

We study whether expectations of order statistics in conjunction with statistical inference can forecast final team places when certain intricate, educated guesses are made about the underlying random process. Specifically, we assume lognormality of both individual leg times and, more importantly, team changeover times.

Generally, in competitive sports, the final place is the hard, quantitative result that both individuals and teams want to minimize. For our case study, both plots and numerical prediction error values show that ordinal regression based on lognormal order statistics of time duration can provide an accurate fit to real-world relay race data.

%% file: method.tex
\section{MODELS}
\subsection{Relay Race System Model}
Consider an orienteering relay race. Let $n$ denote the number of finishing teams as we ignore disqualified and retired teams. Each team has $m$ runners and each runner runs one \emph{leg}. Especially note that $m$ is given, whereas $n$ is estimated, as will be discussed later.

We say that leg time results correspond to random variables $\{Z_i\}$ with $i\in\{1,2,\dots,m\}$. We define the \emph{changeover time} $T^{(l)}$ after leg $l\in\{1,2,\dots,m\}$ as
\begin{align}\label{Tl}
T^{(l)} \coloneqq \sum_{i=1}^l Z_i.
\end{align}
The \emph{order statistics} $T_{r:n}^{(l)}$ of the sum\footnote{The order statistics of addends of ordered sums are latent order statistics, examples of which include factor distributions \cite{JP2019}. In this work, though, we do not need to derive any sum or addend distributions.} $T^{(l)}$ satisfy
\begin{align*}
T_{1:n}^{(l)} \leq T_{2:n}^{(l)} \leq \dots \leq T_{r:n}^{(l)} \leq \dots \leq T_{n:n}^{(l)}.
\end{align*}
The final team result list of the relay race is a length-$n$ sample of $T^{(m)}$ sorted in ascending order.

Let $F_{T^{(m)}}(\cdot)$ denote the cdf of $T^{(m)}$, and $T_{r_m:n}^{(m)}$ denote the $r_m^{\text{th}}$ order statistic of a length-$n$ sample of $T^{(m)}$. These order statistics satisfy $T_{1:n}^{(m)} \leq T_{2:n}^{(m)} \leq \dots \leq T_{r_m:n}^{(m)} \leq \dots \leq T_{n:n}^{(m)}$. We refer to $r_m$ as the (final) \emph{place} of the corresponding team, where $r_m=1$ corresponds to the winning team.

Let $c$ denote the number of training observations. We are given a changeover time training vector $\x_l=\left[x_1^{(l)},\dots,x_c^{(l)}\right]$ and a final place training vector $\y=[y_1,\dots,y_c]$. Hence, $\x_l$ consists of realizations of $T^{(l)}$ and $\y$ includes the corresponding observed places. We wish to find regression functions $h^{(l)}(\cdot)$ that satisfy
\begin{align}\label{yhx}
\y\approx h^{(l)}{\left(\x_l\right)}
\end{align}
as accurately as possible. In \eqref{yhx}, we use vector notation to emphasize that we find the parameters of $h^{(l)}(\cdot)$ that minimize a loss function with a set of observation pairs rather than with a single observation pair.

\subsection{The Four Regression Models}
\paragraph{1) Linear} regression refers here to Ordinary Least Squares (OLS) regression rounded to the nearest integer. OLS finds the intercept and slope that minimize the residual sum of squares between the observed targets and the targets predicted by the linear approximation.

\paragraph{2) Gaussian Process} (GP) regression is a nonparametric model that can manage exact regression up to a million data points on commodity hardware \cite{Wang2019}.

For a pair of training vectors $(\x_l,\y)$, a GP is defined by its \emph{kernel} function $k(\cdot,\cdot)$, a $c{\times}c$ kernel matrix $\K_{\x_l \! \x_l}$ with covariance values for all training point pairs, and a $c$-dimensional vector $\bk_{\x_l x}$ with evaluations of the kernel function between training points $\x_l$ and a given point $x$.

A GP predicts an unknown function $g(\cdot)$. For kernel matrix $\widehat \K_{\x_l \! \x_l} = \K_{\x_l \! \x_l} + \sigma_0^2 I$ with additive Gaussian noise with zero mean and variance $\sigma_0^2$, the expected value of the zero mean GP predictive posterior distribution with a Gaussian likelihood is \cite{Rasmussen2006}
\begin{align}\label{0}
\mathbb{E}\left(g(x) \! \mid \! \x_l, \by\right) = \bk_{\x_l x}\transpose\widehat \K_{\x_l \! \x_l}^{-1}\y.
\end{align}
We use \eqref{0} rounded to the nearest integer as the GP place predictor. The GP regression function thus becomes
\begin{align}\label{00}
h_{\text{GP}}^{(l)}(x) \coloneqq \big[\bk_{\x_l x}\transpose\widehat \K_{\x_l \! \x_l}^{-1}\y].
\end{align}
For practical, numerical implementation of exact GP, we utilize the readily available GPyTorch library with a radial basis function (RBF) kernel as in the ``GPyTorch Regression Tutorial" on \cite{GPytorch}.

\paragraph{3) Ordinal} regression refers here to the rounded to the nearest integer regression-based model from the readily available Python mord package for ordered ordinal ridge regression. This model overwrites the ridge regression function from the scikit-learn library and uses the (minus) absolute error as its score function \cite{Mord, Fabian2015}.

\paragraph{4) Fenton-Wilkinson Order Statistics} (FWOS) regression is our original regression model. For this model we make the following two well-educated assumptions.

\textbf{Assumption 1:} Individual leg time $Z_i$ is lognormal.

\textbf{Assumption 2:} Changeover time $T^{(l)}$ is lognormal.

The lognormal distribution often appears in sciences \cite{Limpert2001}. Assumption 1 is based on the lognormality of vehicle travel time \cite{Chen2018}.

Assumption 2 paraphrases what in the literature is known as the \emph{Fenton-Wilkinson approximation}
\cite{Wilk1967,Fenton1960,Barry2012}. The Fenton-Wilkinson approximation method is the method of approximating the distribution of the sum of lognormal random variables with another lognormal distribution.

Note that $Z_i$ and $T^{(l)}$ are both lognormal and independent but not identically distributed. With this in mind, we can now derive the FWOS regression prediction function $h_{\text{OS}}^{(l)}(\cdot)$. For this purpose we use the following two well-known preliminary tools in probability theory.

\textbf{Tool 1:} Let $W$ follow the standard uniform distribution $U(0,1)$ and $T^{(l)}$ follow distribution $F$. Let $W_{r:n}$ denote the $r^{\text{th}}$ order statistic of a length-$n$ sample of $W$. The $r^{\text{th}}$ order statistic of a length-$n$ sample of $T^{(l)}$ has the same distribution as the inverse cdf of $F$ at $W_{r:n}$

\textbf{Tool 2:} The $r^{\text{th}}$ standard uniform order statistic follows Beta($r,n - r + 1$). Therefore, $\mathbb{E}(W_{r:n}) = r/(n+1)$.

The inverse cdf of $F$ is known as the \emph{quantile function} $Q_{F}(\cdot)$. Tool 1 can be therefore expressed as
\begin{align}\label{Trn}
T_{r:n}^{(l)} \stackrel{\text{d}}{=} Q_{F}(W_{r:n}),
\end{align}
where ``$\stackrel{\text{d}}{=}$" reads ``has the same distribution as", and applying Tool 2 to \eqref{Trn} hence yields
\begin{align}\label{ETrn}
\mathbb{E}(T_{r:n}^{(l)}) = Q_{F}{\left(\frac{r}{n+1}\right)}.
\end{align}
Let $F$ be the lognormal distribution with cdf $F_{T^{(l)}}(\cdot)$. Now \eqref{ETrn} directly implies $F_{T^{(l)}}{\left(\mathbb{E}(T_{r:n}^{(l)})\right)} = r/(n + 1)$ and, further, most interestingly for our purposes, that
\begin{align}\label{1}
F_{T^{(l)}}{\left(\mathbb{E}(T_{r:n}^{(l)})\right)}(n + 1) = r,
\end{align}
which resembles \eqref{yhx} as $r$ is associated with $y$.

For large $n$, as in our case study, it is reasonable to assume that $\forall x\in\mathbb{R}_+, \exists r$ such that
\begin{align}\label{2}
\mathbb{E}\left(T_{r:n}^{(l)}\right)\approx x.
\end{align}

The well-known lognormal cdf is given by
\begin{align}\label{3}
F_{T^{(l)}}{(x;\mu_l,\sigma_l)} = \Phi\left(\frac{\log x - \mu_l}{\sigma_l}\right),
\end{align}
where $\Phi(\cdot)$ is the standard normal cdf, $\mu_l$ and $\sigma_l$ are the lognormal parameters and $\log(\cdot)$ is the natural logarithm.

We combine \eqref{1}, \eqref{2} and \eqref{3}, and define the FWOS place prediction function for leg $l$ as
\begin{align}\label{4}
h_{\text{OS}}^{(l)}(x) \coloneqq \Bigg[\Phi\left(\frac{\log x - \mu_l}{\sigma_l}\right)(n + 1)\Bigg],
\end{align}
where $[\cdot]$ again denotes rounding to the nearest integer.

Maximum likelihood estimation (MLE) for the normal distribution yields lognormal estimators for $\mu_l$ and $\sigma_l$ when $x_i^{(l)}$ are replaced by $q_i^{(l)}\coloneqq\log x_i^{(l)}$. Hence,
\begin{align}\label{6}
\left(\hat{\mu_l},\hat{\sigma_l}^2\right)=\left(\frac1c\sum_{i=1}^cq_i^{(l)},\frac1c\sum_{i=1}^c \left(q_i^{(l)} - \hat{\mu_l}\right)^2\right).
\end{align}

What remains to be done is finding an estimate for the total number of teams $n$. We assume that there are no ties, which is equivalent to stating that the elements in $\y$ are unique. Thus, $\y$ is a sample, without replacement, of the discrete uniform distribution $\mathcal{U}[1,n]$. Let $D$ denote a random variable that follows this distribution. Now recall that $\y$ is a random $c$-subset of $\{1,2,\dots,n\}$.

Estimating the parameter $n$ of $\mathcal{U}[1,n]$ with a sample drawn without replacement is in the literature known as the \emph{German tank problem} \cite{Ruggles1947}, a solution to which is a uniformly minimum-variance unbiased estimator (UMVUE) \cite{Goodman1952}
\begin{align}\label{7}
\hat{n} = \left(1+\frac1c\right)d_{(c)} - 1,
\end{align}
where $$d_{(c)}\coloneqq\max_{i\in\{1,2,\dots,c\}} y_i$$ is the realization of the $c^\text{th}$ order statistic (maximum) of a length-$c$ sample of $D$.

We replace $n$ and $(\mu_l,\sigma_l)$ in \eqref{4} with the $\hat{n}$ of \eqref{7} and $\left(\hat{\mu_l},\hat{\sigma_l}\right)$, respectively, and note that numerical values for $\left(\hat{\mu_l},\hat{\sigma_l}\right)$ are obtained through \eqref{6}. We obtain the following proposition.

\begin{prop}
For a training dataset of $c$ observations, lognormal parameter estimates $(\hat{\mu}_l,\hat{\sigma}_l)$ that are obtained through the maximum likelihood estimators of changeover time observations $\x_l$, and $d_{(c)}$, the maximum of final place observations $\y$, the FWOS ordinal regression function
\begin{align*}
h_{\text{OS}}^{(l)}(x) = \Bigg[\Phi\left(\frac{\log x - \hat{\mu}_l}{\hat{\sigma}_l}\right)\left(1+\frac1c\right)d_{(c)}\Bigg]
\end{align*}

predicts final place with changeover time $x\in\mathbb{R}_+$ at changeover $l$.
\end{prop}

Loosely speaking, Proposition 1 states that the FWOS method approximates final place with expected changeover place. We anticipate that this approximation holds to a satisfactory degree and that it improves with $l$.

We note that $c$, the dimension of the training vectors, is constant for all changeovers $l$ as is $\hat{n}$, the estimate of $n$, while the pair $(\mu_l,\sigma_l)$ is separately estimated for each changeover $l$ with changeover time training vector $\x_l$.

Further note that FWOS regression only requires an easily acquirable pair $\left(c,d_{(c)}\right)$ of dimension $2$ as opposed to the other three regression methods that require the whole place training vector $\y$ of dimension $c$.

%% file: results.tex
\section{NUMERICAL RESULTS}
Real-world data are acquired from the results of Jukola 2019 \cite{Jukola2019} with $n=1653$ teams with $m=7$ runners per team. Training of the regression models is conducted for two cases: for $c=1322$ $\left(c/n\approx80\%\right)$ and also for $c=82$ $\left(c/n\approx5\%\right)$ training observations. The rest of the data are used for testing the models.

\begin{figure*}[t!]
  \centering
  \includegraphics[width=.87\linewidth]{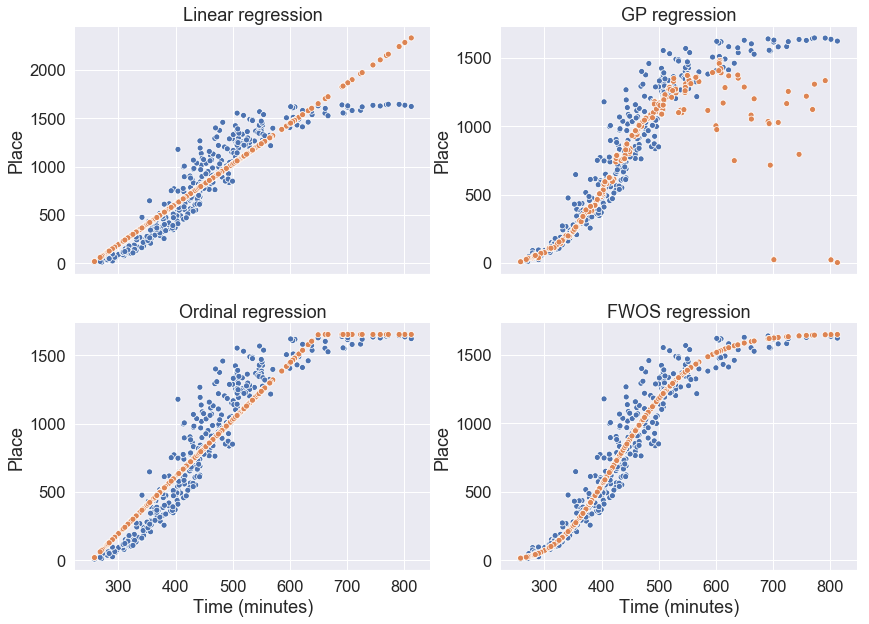}
  \caption{Place predictions (orange) at changeover $l=4$ with training set size 80\% and test set (blue) size 20\%.}
  \label{fig:leg4ts20}
\end{figure*}

\begin{figure*}[t!]
  \centering
  \includegraphics[width=.87\linewidth]{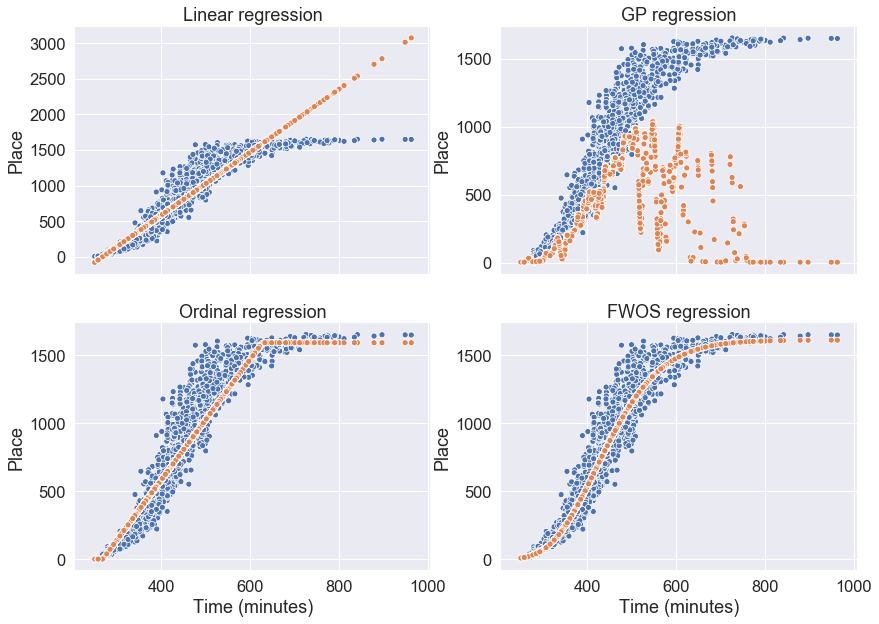}
  \caption{Place predictions (orange) at changeover $l=4$ with training set size 5\% and test set (blue) size 95\%.}
  \label{fig:leg4ts95}
\end{figure*}

\begin{figure*}[t!]
\centering
\begin{subfigure}{.49\textwidth}
  \centering
  \includegraphics[width=1\linewidth]{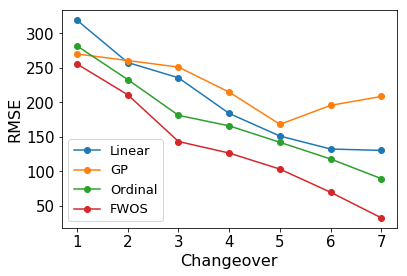}
  \caption{Training set size $80\%$.}
  \label{fig:rmse20}
\end{subfigure}
\begin{subfigure}{.49\textwidth}
  \centering
  \includegraphics[width=1\linewidth]{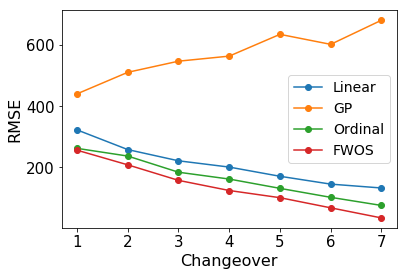}
  \caption{Training set size $5\%$}
  \label{fig:rmse95}
\end{subfigure}
\caption{Place prediction root-mean-square errors (RMSEs) after each leg.}
\label{fig:test}
\end{figure*}

\subsection{Regression Curves}
Figure \ref{fig:leg4ts20} plots place against team changeover time after leg $l=4$ for the predictions (orange) and the test set (blue) for all the four regression models. The size of the training set is $80\%$ and the size of the test set is $20\%$.

We see that linear regression and GP regression provide accurate fits for a large portion of the test set points, but behave poorly when time is large. For average teams, with approximately $350$ to $550$ minutes of elapsed time after four legs, place seems to grow linearly with time, though overall linearity is clearly an oversimplification.

We further notice that ordinal regression captures the effect where place saturates for large values of time, but fails to provide a smooth transition. FWOS regression, unlike the other models, does indeed exhibit the smooth sigmoidal behavior of the data.

In the setting of Fig. \ref{fig:leg4ts95}, there are significantly fewer training data compared to the setting of Fig. \ref{fig:leg4ts20}, namely, $5\%$ compared to $80\%$. Interestingly, linear regression, ordinal regression and FWOS regression maintain high performance, whereas GP regression greatly suffers from the lack of training data.

Overall, we make the following two key remarks.

\textbf{Remark 1:} The ``place against changeover time" curve resembles a scaled lognormal cumulative distribution function. This further suggests that elite teams ``pull away'' from the rest of the teams, while extremely slow teams fall far behind the rest.

\textbf{Remark 2:} A lognormal cumulative distribution function, alongside a German tank estimator, effectively predicts places with changeover times even when the number of training observations is small.

\subsection{Root-Mean-Square Errors}
Here we illustrate the error between place prediction $h^{(l)}(x_i)$ and the corresponding true value $y_i$ with label $i\in\{1,2,\dots,v\}$, where $v$ is the size of the test set with $n=c+v$. We use the root-mean-square error (RMSE)

\begin{align*}
\text{RMSE} &= \sqrt{\frac{1}{v}\sum_{i=1}^{v} \big(y_i - h^{(l)}(x_i)\big)^2}.
\end{align*}

Fig. \ref{fig:rmse20} plots the RMSEs after each of the $m=7$ changeovers for a random test set of $v=331$ points ($v/n\approx0.20$ as in Fig. \ref{fig:leg4ts20}), while Fig. \ref{fig:rmse95} plots the RMSEs for a random test set of $v=1571$ points ($v/n\approx0.95$ as in Fig. \ref{fig:leg4ts95}).

When comparing Fig. \ref{fig:rmse20} with Fig. \ref{fig:rmse95}, we notice similar RMSEs for both training set sizes, except for GP regression. It is clear that GP regression requires more training data than the other regression models to achieve comparable prediction error performance.

In every case, FWOS exhibits the best RMSE performance. However, the RMSEs are not zero even for the $7^{\text{th}}$ changeover, \emph{i.e.}, after the \emph{anchor leg} when the team finishes. This is due to the imperfections of the regression models and the random fluctuations of the data.

%% file: discussion.tex
\section{DISCUSSION}
For our case study of an orienteering relay race, we have shown that rankings of ordered sums, here referred to as \emph{places}, can be relatively accurately predicted by intermediate sums by first taking an educated guess that leg times are lognormal, and then by scaling the cumulative distribution function of another lognormal distribution corresponding to observed changeover times. The latter distribution can be approximated by the Fenton-Wilkinson method, while the scaling factor can be estimated with a well-known solution to the German tank problem.

The use of order statistics for ordinal regression provides a powerful approximation for our case study. However, intricate prior insight into the underlying random process is assumed, which is unnecessary for nonparametric models such as the Gaussian process. Similarly, linear regression and standard ordinal regression require no knowledge about the underlying distributions.

An event that may complicate leg time distribution fitting and erode our prediction performance is the \emph{restart}, where the runners that have not started before a specific cut-off time start simultaneously. Restarts may skew the leg time distributions in the same way as the first leg runners that often run in packs. Running in packs may decrease our prediction accuracy.

Fenton-Wilkinson order statistics, alongside German tank estimators, could be applied whenever a longer random process can be modeled as a sum of shorter random processes with random durations and when the outcomes of the longer process are uniquely ordered by the corresponding total time duration. Such processes are plentiful in sports -- those sources of inspiration for ordinal classification order statistics research.